\documentclass[10pt,twocolumn,letterpaper]{article}

\usepackage{cvpr}      

\usepackage{graphicx}
\usepackage{amsmath}
\usepackage{amssymb}
\usepackage{booktabs}
\usepackage[accsupp]{axessibility}
\usepackage{enumitem}

\setitemize[1]{itemsep=0pt,partopsep=0pt,parsep=\parskip,topsep=3pt}

\newcommand{\YL}[1]{{\color{black}#1}}
\newcommand{\HY}[1]{{\color{black}#1}}

\usepackage[pagebackref,breaklinks,colorlinks]{hyperref}

\usepackage[capitalize]{cleveref}
\crefname{section}{Sec.}{Secs.}
\Crefname{section}{Section}{Sections}
\Crefname{table}{Table}{Tables}
\crefname{table}{Tab.}{Tabs.}

\def\cvprPaperID{2764} 
\def\confName{CVPR}
\def\confYear{2023}

\begin{document}

\title{Towards Artistic Image Aesthetics Assessment: \\ \YL{a} Large-scale Dataset and \YL{a} New Method}

\author{Ran Yi$^{1}$\thanks{Corresponding author.}, Haoyuan Tian$^{1}$, Zhihao Gu$^{1}$, Yu-Kun Lai$^{2}$, Paul L. Rosin$^{2}$\\
$^{1}$Shanghai Jiao Tong University, $^{2}$Cardiff University\\
{\tt\small \{ranyi,thy0210,ellery-holmes\}@sjtu.edu.cn, \{LaiY4,RosinPL\}@cardiff.ac.uk}
}
\maketitle

\begin{abstract}
  Image aesthetics assessment (IAA) is a challenging task due to its highly subjective nature. Most of the current studies rely on large-scale datasets (e.g., AVA and AADB) to learn a general model for all kinds of photography images. However, little light has been shed on measuring the aesthetic quality of artistic images, and the existing datasets only contain relatively few artworks. Such a defect is a great obstacle to the aesthetic assessment of artistic images.
  To fill the gap in the field of artistic image aesthetics assessment (AIAA), we first introduce a large-scale AIAA dataset: Boldbrush Artistic Image Dataset (BAID), which consists of 60,337 artistic images covering various art forms, with more than 360,000 votes from online users. We then propose a new method, SAAN (Style-specific Art Assessment Network), which can effectively extract and utilize style-specific and generic aesthetic information to evaluate artistic images. Experiments demonstrate that our proposed approach outperforms existing IAA methods on the proposed BAID dataset according to quantitative comparisons. We believe the proposed dataset and method can serve as a foundation for future AIAA works and inspire more research in this field. Dataset and code are available at: \url{https://github.com/Dreemurr-T/BAID.git}
\end{abstract}

\section{Introduction}
\label{sec:intro}

With the ever-growing scale of online visual data, image aesthetic assessment (IAA) shows great potential in a variety of applications such as photo recommendation, image ranking and  image search~\cite{deng2017image}. In recent years, image style transfer \cite{gatys2016image, huang2017arbitrary, park2019arbitrary, li2017universal, liu2021adaattn} and AI painting \cite{zhang2018ai, huang2019learning} have become high-profile research areas. Users can easily generate artworks of numerous styles from websites and online applications, which has led to the explosion of artistic images online and the drastic increase in demand for automatically evaluating
artwork aesthetics. We refer to this problem as \textbf{artistic image aesthetic assessment (AIAA)}.

\begin{figure}[t]
  \centering
   \includegraphics[width=0.99\linewidth]{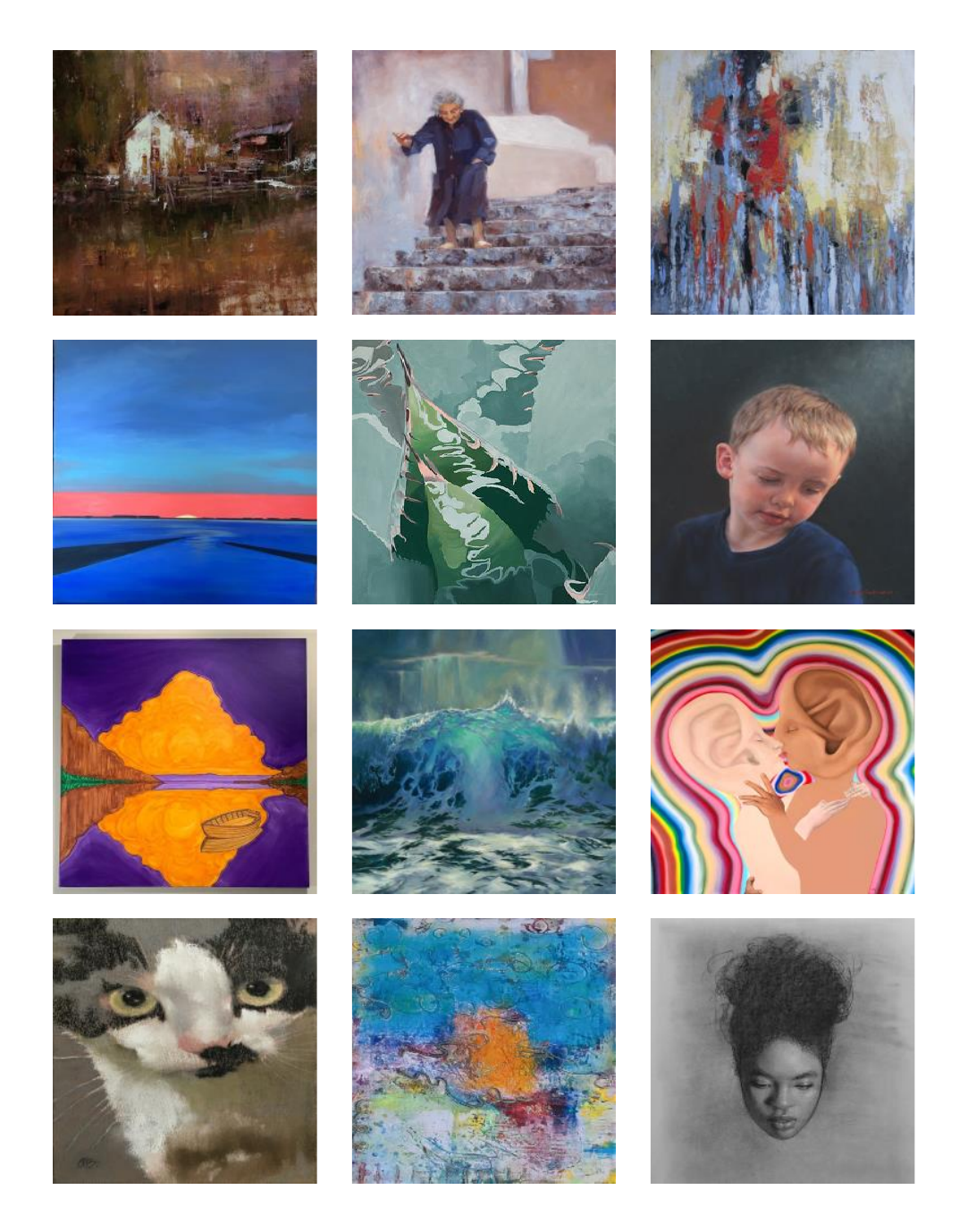}
   \caption{Samples from the proposed BAID \YL{dataset}. BAID covers a wide range of artistic styles and painting themes.}
   \label{fig:images}
\end{figure}

The artistic image aesthetic assessment task is similar to IAA for being extremely challenging \YL{due to its} highly subjective \YL{nature}, as different individuals may have distinct visual and art preferences. Existing datasets related to this task can be summarized into three categories, but none of them meets the requirements of the AIAA task:
(1) {\bf IAA datasets}: modern IAA methods \cite{lu2015rating, ma2017lamp, talebi2018nima, hosu2019effective, sheng2018attention, she2021hierarchical} are data-driven, usually trained and evaluated on large-scale IAA datasets, \eg, AVA \cite{murray2012ava}, AADB \cite{kong2016photo} and CUHK-PQ \cite{luo2011content}. However, these datasets only contain real-world photos and do not include artistic images like oil paintings or pencil sketches. This deficiency of artistic images is prevalent in existing IAA datasets \cite{datta2008algorithmic, joshi2011aesthetics, luo2011content, kong2016photo, ren2017personalized}, which means that given an artwork, existing IAA methods evaluate it based on perceptions learned from photography, and the evaluation is likely to be inaccurate since the perceptual rules of photography and art are not the same. 
(2) {\bf Artistic datasets without aesthetic labels}: existing large-scale artistic image datasets \cite{wilber2017bam, saleh2015large, achlioptas2021artemis} are mainly used to train style transfer, artistic style classification or text to image models, but they lack score annotations indicating image aesthetic level. 
(3) {\bf Small-scale AIAA datasets}: efforts into building public AIAA datasets are scarce and the existing datasets~\cite{amirshahi2014jenaesthetics, fekete2022vienna} contain relatively few number of images (less than 2,000). Based on the above observations, we conclude that {\bf the lack of a large-scale AIAA dataset} is the biggest obstacle towards developing AIAA approaches. 

To solve the problem, we first introduce a large-scale dataset specifically constructed for the AIAA task: the Boldbrush Artistic Image Dataset (BAID), which consists of 60,337 artworks annotated with more than 360,000 votes. The proposed BAID is, to our knowledge, the largest AIAA dataset, 
which far exceeds existing IAA and AIAA datasets in the quantity and quality of artworks.

Furthermore, we propose a baseline model, called the Style-specific Art Assessment Network (SAAN), which can effectively exploit the style features and the generic aesthetic features of the given artwork. Our model consists of three modules: 1) {\bf Generic Aesthetic Feature Extraction Branch: }inspired by the studies \cite{sheng2020revisiting, Pfister_2021_CVPR}, we adopt a self-supervised learning scheme to train a Generic Aesthetic Branch to extract aesthetics-aware features. The self-supervised scheme is based on the correlation between the aesthetic quality of the images and degradation editing operations. 
\YL{This essentially provides data augmentation such that the model can better learn the quality of different artworks.}
2) {\bf Style-specific Aesthetic Feature Extraction Branch: }observing that the style of the artwork is critical when assessing its aesthetic value and different styles need to extract different style-related aesthetic features, we propose a Style-specific Aesthetic Branch to incorporate style information into aesthetic features and extract style-specific aesthetic features via adaptive instance normalization \cite{huang2017arbitrary}. 3) {\bf Spatial Information Fusion:} we also add a non-local block \cite{wang2018non} into the proposed method to fuse spatial information into the extracted aesthetic features.

The main contributions of our work are three-fold:
\begin{itemize}
    \item We address the problem of artistic image aesthetics assessment, and introduce a new large-scale dataset BAID consisting of 60,337 artworks annotated with more than 360,000 votes to facilitate research in this direction.
    \item We propose a style-specific artistic image assessment network called SAAN, which combines style-specific and generic aesthetic features to evaluate artworks.
    \item We evaluate the state-of-the-art IAA approaches and our proposed method on the proposed BAID dataset. Our model achieves promising results on all the metrics, which clearly demonstrates the validity of our model.

\end{itemize}

\begin{table}[t]
\caption{Summary of IAA/AIAA datasets and our proposed BAID dataset. BAID provides a significantly larger number of artistic images and has user subjective votes.}
\resizebox{.95\columnwidth}{!}{

\begin{tabular}{ccc}
\toprule[2pt]
\textbf{Dataset} & \textbf{Number of images} & \textbf{Number of {\bf artistic} images} \\ \midrule[1pt]
DP Challenge \cite{datta2008algorithmic}     & 16,509                   & --                                 \\
Photo.Net \cite{joshi2011aesthetics}       & 20,278                   & --                                 \\
CUHK-PQ \cite{luo2011content}         & 17,673                   & --                                 \\
AVA \cite{murray2012ava}         & 255,530                  & --                                 \\
AADB    \cite{kong2016photo}  & 10,000                   & --                                 \\
FLICKR-AES    \cite{ren2017personalized}   & 40,000                   & --                                 \\
PARA    \cite{yang2022personalized}         & 31,220                   & --                                 \\
TAD66K  \cite{ijcai2022p132}         & 66,327                   & 1,200                             \\
JenAesthetic \cite{amirshahi2014jenaesthetics} & 1,628 & 1,628\\
VAPS \cite{fekete2022vienna} & 999 & 999\\
\midrule[1pt]
BAID (Ours)      & 60,337                   & \textbf{60,337}                            \\
\bottomrule[2pt]
\end{tabular}
}

\label{tab:datasets}
\end{table}

\section{Related Work}
\noindent
\textbf{Image Aesthetic Assessment Datasets.} The Photo.net dataset \cite{joshi2011aesthetics} and the DPChallenge dataset \cite{datta2008algorithmic} are the earliest attempts to construct public image databases for IAA. The Chinese University of Hong Kong-Photo Quality (CUHK-PQ) dataset is introduced in \cite{luo2011content}, which is the first dataset organized by topics. The AVA dataset\cite{murray2012ava} consists of approximately 255,000 images derived from DPChallenge.com with aesthetic annotations. Additionally, the AVA dataset contains photographic-style attributes and category attributes for a subset of images. Kong \etal ~\cite{kong2016photo} provided a new dataset called the Aesthetics and Attributes Database (AADB), which includes individual ratings of aesthetics and attributes of multiple images. Ren \etal \cite{ren2017personalized} and Yang \etal \cite{yang2022personalized} constructed FLICKR-AES and PARA respectively for personalized image aesthetic assessment. He \etal \cite{ijcai2022p132} introduced a theme-oriented dataset TAD66K which includes 47 themes and a unique \YL{criterion} 
for each specific theme.

Although the above datasets have provided a solid foundation for IAA methods, they rarely include art images and consider different evaluation criteria for photos and artworks. As for existing AIAA datasets, neither of the public datasets Jenaesthetics \cite{amirshahi2014jenaesthetics} (1,628 art images) or VAPS (Vienna Art Picture System) \cite{fekete2022vienna} (999 paintings) is large enough to meet the requirements of deep learning methods.

In contrast, we construct the BAID dataset, which is, to our knowledge, the largest AIAA dataset made up entirely of artistic images (60,337 in total) and densely annotated with scores (more than 360,000 votes). The comparison of our BAID and the existing datasets is listed in \cref{tab:datasets}.

\noindent \textbf{Image Aesthetic Assessment Models.} Early studies on IAA mainly focus on designing and extracting handcrafted features from images and mapping the features to annotated aesthetics labels \cite{dhar2011high, marchesotti2011assessing}. With the emergence of large-scale IAA datasets \cite{murray2012ava, kong2016photo}, methods based on deep learning continue to develop. NIMA \cite{talebi2018nima} utilized Earth Mover's Distance (EMD) loss to predict the distribution of aesthetic scores. ${\rm MP}_{ada}$ \cite{sheng2018attention} adopts an attention-based mechanism to dynamically adjust the weights of each patch during the training process to improve learning efficiency. Hosu \etal \cite{hosu2019effective} propose the first AIAA method that efficiently supports full resolution images as an input, and can be trained on variable input sizes. \cite{ma2017lamp} uses a saliency detection model to extract some more representative image patches, which are then fed into the network to extract features. She \etal \cite{she2021hierarchical} present a Hierarchical Layout-Aware Graph Convolutional Network (HLA-GCN) to capture layout information. TANet \cite{ijcai2022p132} can adaptively learn the rules for predicting aesthetics according to a recognized theme.

There are relatively few AIAA methods, 
where earlier traditional methods \cite{6738080, 4799314, 8227452} design handcrafted features and train Support Vector Machine (SVM) for classification. Recently Zhang \etal \cite{9293299} 
developed a deep multi-view parallel convolutional neural network (DMVCNN) 
to learn aesthetic features for Chinese ink paintings. 
In general, AIAA methods have not been adequately studied.

Different from the above works, we argue that different art styles need to extract different style-related aesthetic features, and combine both style-specific and generic aesthetic features to evaluate artworks.

\section{Boldbrush Artistic Image Dataset}
In this section, we discuss the data collection and the generation of scores of the proposed BAID dataset.

\subsection{Data Collection}
Constructing an artistic image dataset with score annotations is arduous. Most online art communities and professional artistic websites do not have public channels to score for artworks since the aesthetics of artworks are quite subjective and the scoring format is somewhat disrespectful to the artists, which has led, to some extent, to the inadequacy of the existing dataset.

We chose to use the website Boldbrush\footnote{https://faso.com/boldbrush/popular} as the source of data. Boldbrush hosts a monthly artwork contest where certified artists can upload their works and receive public votes from online users. 
Users can click into the detail page of the artwork and vote for it if they like the artwork, which means that the more votes, the greater the number of people consider the artwork pleasing and good-looking. Note that the users can vote for as many artworks as they like, and their individual votes are not ranked.

The benefits of our choice are as follows:
\begin{itemize}
    \item The competition does not limit the subject matter, style or medium used to create the work, thus the website contains artworks with various art styles and contents.
    \item Every time a voter wants to place a vote, he/she will receive a verification email 
    to confirm the vote. Moreover, the website performs email address check to prevent users from voting for the same work more than once. Thus, the votes will not suffer from malicious vote fraud \HY{and are more reliable than the `favourite' annotations on other art communities like Flickr\footnote{https://www.flickr.com/}}. 
    \item Boldbrush and the FASO organization have a high profile in the art world. They have been holding such contests since July 2010. \HY{The voters are largely made up of artists and art collectors,} so the results have a high degree of credibility and authority.
\end{itemize}

A total of 60,408 images and the corresponding annotations were collected, and 60,337 images are valid and available after removing corrupted data. Note that we exclude images with 0 vote since they are not included in the popular entries of BoldBrush.

\begin{figure*}[t]
  \centering
  \begin{subfigure}{0.49\linewidth}
    \includegraphics[width=.95\linewidth, height=.5\linewidth]{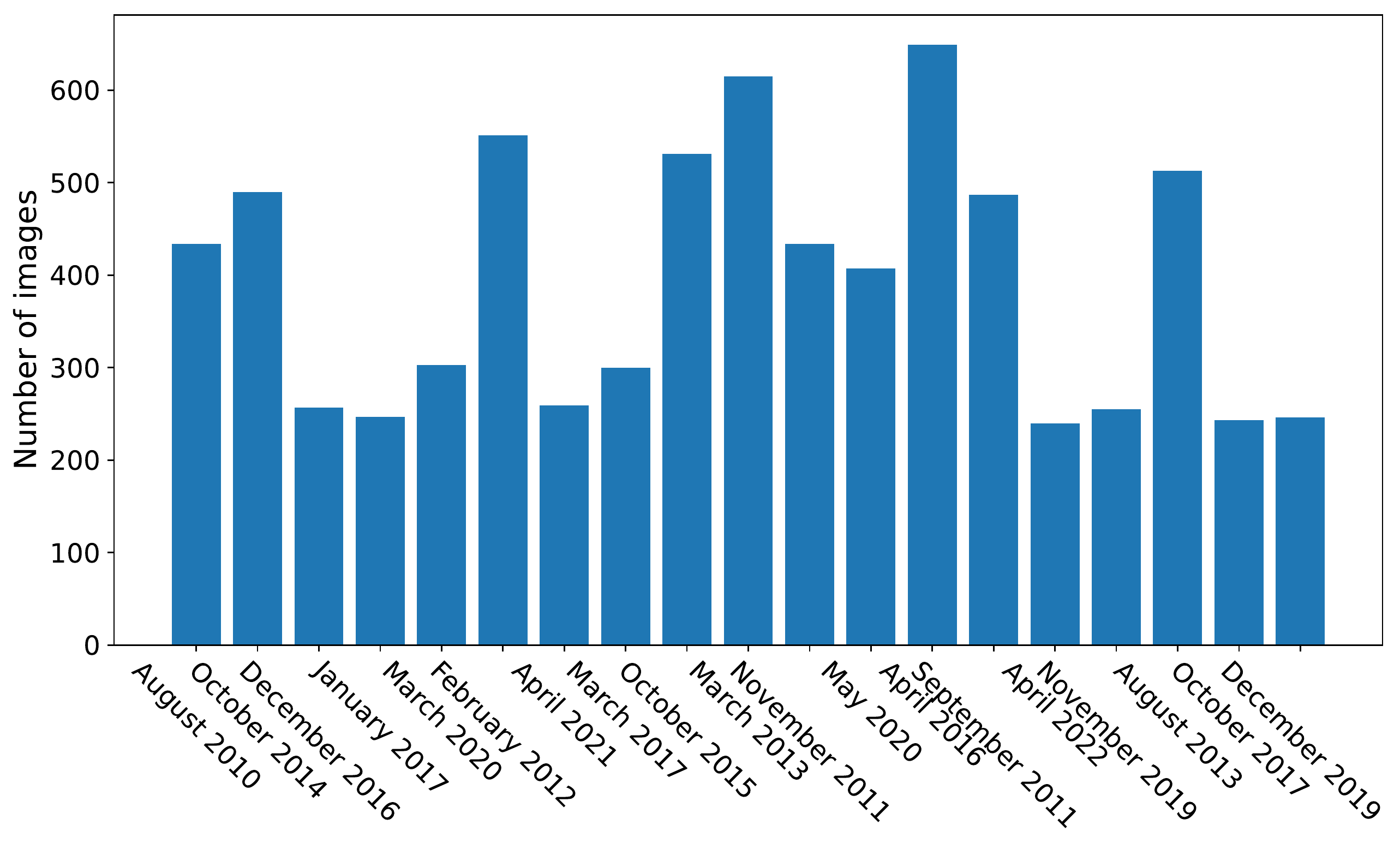}
    \caption{The number of images collected in 20 randomly selected months.}
    \label{fig:numimages}
  \end{subfigure}
  \hfill
  \begin{subfigure}{0.49\linewidth}
    \includegraphics[width=.95\linewidth, height=.5\linewidth]{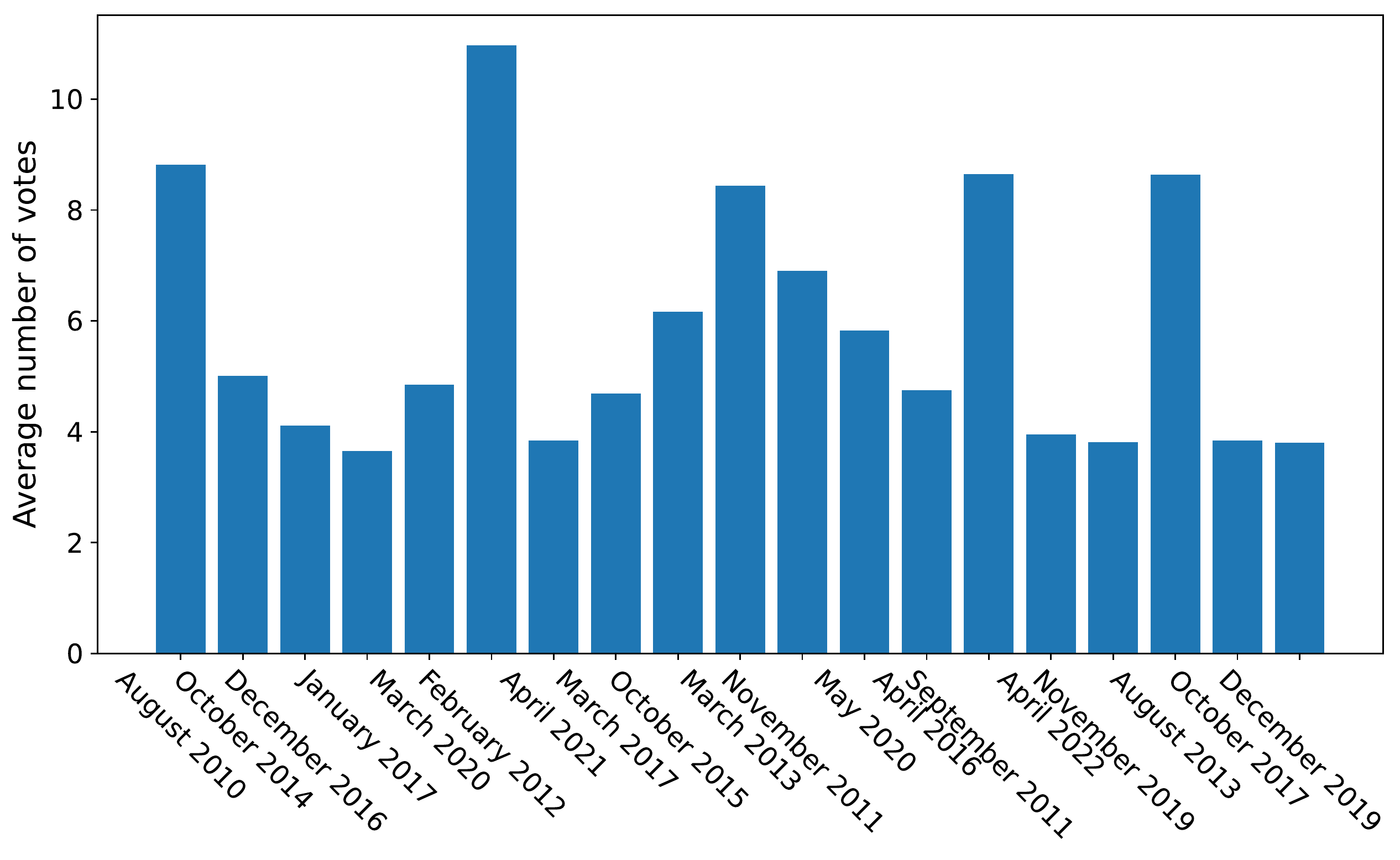}
    \caption{The average number of votes in 20 randomly selected months.}
    \label{fig:avgvote}
  \end{subfigure}
  \begin{subfigure}{0.49\linewidth}
    \includegraphics[width=.95\linewidth]{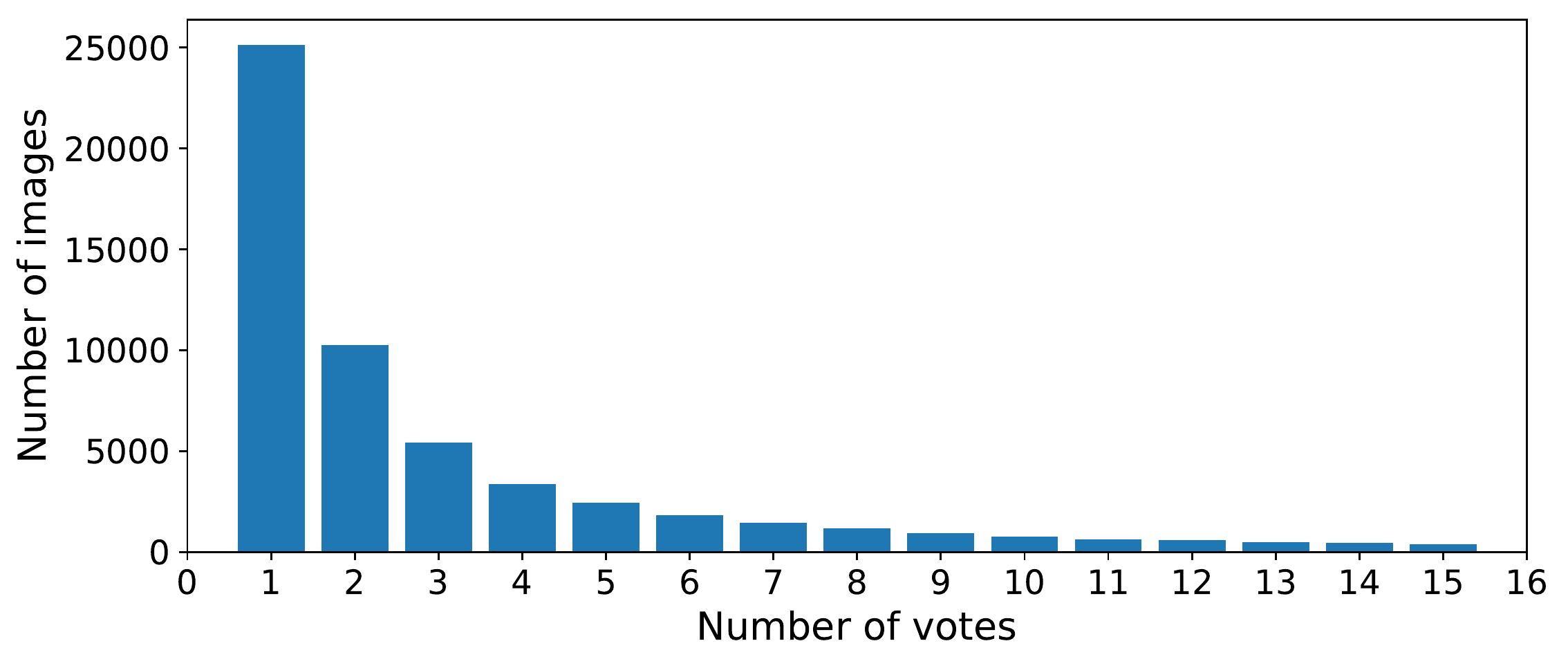}
    \caption{The vote distribution of BAID.}
    \label{fig:vote}
  \end{subfigure}
  \begin{subfigure}{0.49\linewidth}
    \includegraphics[width=.95\linewidth]{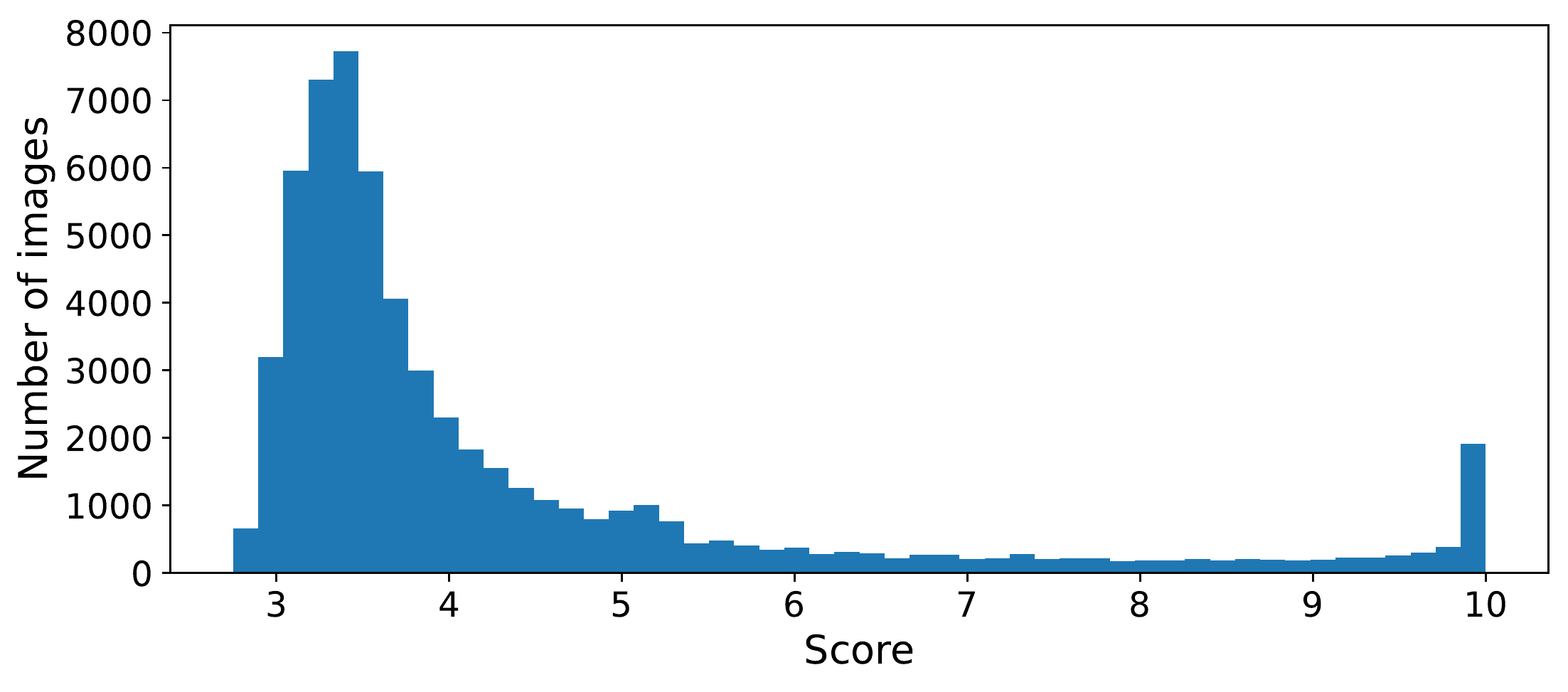}
    \caption{The score distribution of BAID.}
    \label{fig:score}
  \end{subfigure}
  \caption{Statistics of the proposed BAID.}
  \label{fig:stats}
\end{figure*}

\begin{figure*}[t]
  \centering
\includegraphics[width=.95\linewidth]{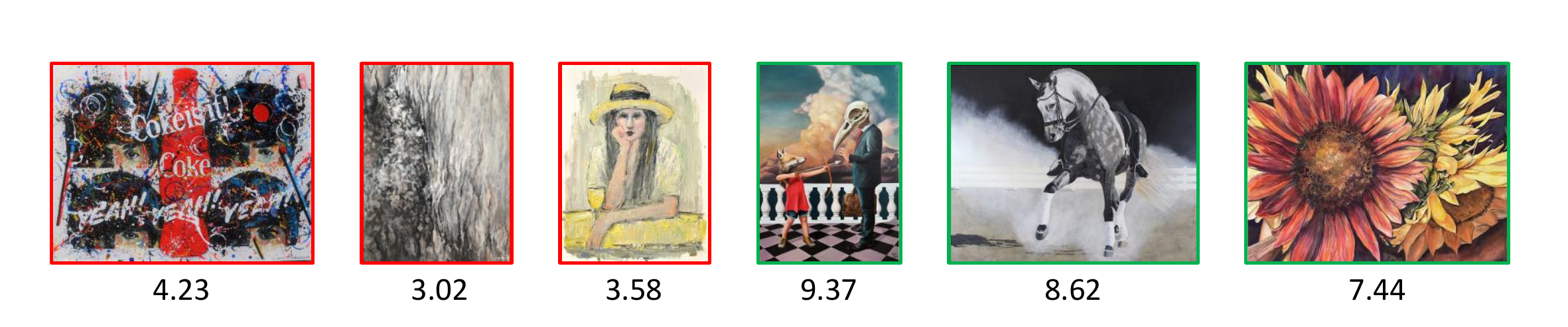}
  \caption{Samples from BAID with generated scores (the number below is the aesthetic score of the image). Low score artworks are marked in red border and high score artworks are marked in green border.}
  \label{fig:samples_score}
\end{figure*}

\subsection{Score Generation}
\label{sec:scoregen}
Unlike the existing IAA datasets where the score distribution is used to calculate the mean opinion score (MOS), we convert the number of votes to the scores of images in BAID. Simply put, images with a higher number of votes are considered to have a higher aesthetic value. Following the common practice, we choose to scale the number of votes into the [0, 10] score range, \YL{where 0 means the worst and 10 means the best.} To elaborate the way we used to generate the scores, two characteristics of the contests' results need to be described:

\begin{itemize}
    \item The number of votes received by entries in 
    a month varies greatly. The margin between the highest number of votes and the lowest over a month can exceed 200.
    \item The images in the proposed dataset are all created by artists with a certain level of skills, thus the overall aesthetic quality is relatively high.
\end{itemize}

The distribution of the number of votes is shown in \cref{fig:vote}. Since the margin between the highest number of votes and the lowest are too large to show in a figure (as described in \cref{sec:scoregen}), we choose images with 1 to 15 votes to demonstrate the overall distribution. Based on the two observations above, using linear mapping from votes to scores is not reasonable since it will make entries with low vote counts receive too low a score. After multiple attempts, we adopt a sigmoid-like way to generate the scores. Specifically, given an image with the number of votes $v_{i}$ and the entry month $m_{i}$, the score $s_{i}$ is calculated using \cref{eq:score}:
\begin{equation}
\begin{aligned}
    x_{i} &= \frac{\bar{v}_{m_{i}} - v_{i}}{\bar{v}_{m_{i}}},\\
    s_{i} &= 10 \times \frac{1}{1+e^{x_{i}}}  ,
\end{aligned}
\label{eq:score}
\end{equation}
where $\bar{v}_{m_{i}}$ is the average number of votes of month $m_{i}$. The final score distribution of the BAID is shown in \cref{fig:score}. Note that the original vote distribution is imbalanced and does not resemble a Gaussian distribution like most IAA datasets \cite{datta2008algorithmic, murray2012ava, joshi2011aesthetics, luo2011content, kong2016photo}: the number of images with low vote counts accounts for a large portion of the proposed dataset. While converting the number of votes to scores, we retain the characteristics of the original distribution due to the high credibility of the data source.

\begin{figure*}[htbp]
    \centering
    \includegraphics[width=.95\linewidth]{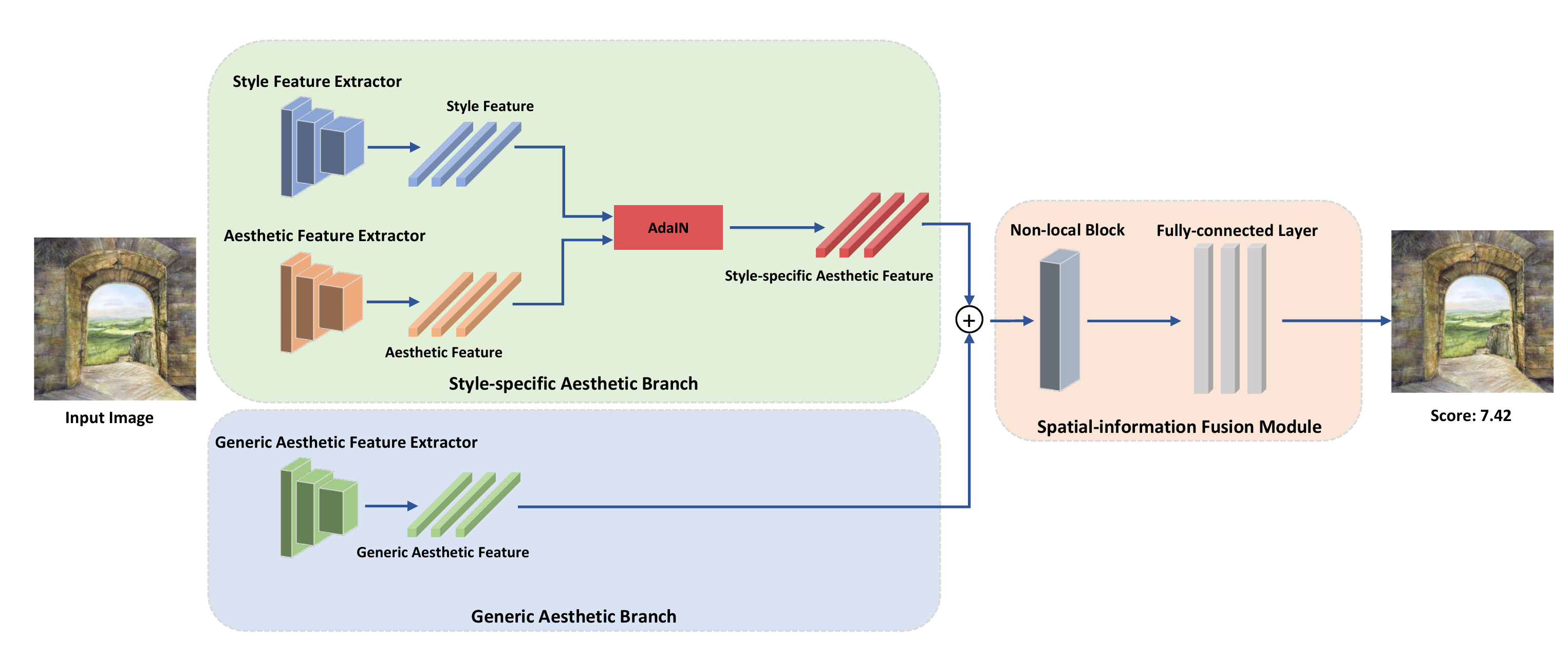}
    \vspace{-0.1in}
    \caption{Overall architecture of the proposed SAAN. SAAN consists of three modules: 1) a style-specific branch to extract style-specific features; 2) a generic aesthetic branch to extract generic aesthetic features; and 3) a spatial information fusion module \YL{that} fuses the spatial information using a non-local block and considers the composition of artwork \YL{during} the assessment. See \cref{sec:SAAN} for further details.}
    \vspace{-0.1in}
    \label{fig:model}
\end{figure*}

\subsection{Further Analysis}
To better demonstrate the data source (BoldBrush) and to support our selected score generating method, we randomly choose 20 months of data from BAID for illustration. \cref{fig:numimages} and \cref{fig:avgvote} show the number of images and the average number of votes of the selected 20 months respectively. \cref{fig:numimages} demonstrates that the data sources are well balanced with the number of entries available each month being above 200, and the gap between months is not too large. \cref{fig:avgvote} indicates that the average number of votes received by each month's entries varies, which means that the average aesthetic quality of the entries and the preference of voters may be different in each month of the contest. Thus, using a fixed threshold to generate binary labels or scores is not reasonable. Here we make use of the average number of votes received by the works in the month for normalization as described in \cref{sec:scoregen}.

\HY{Since we do not have detailed information about the voters, we conducted an MOS (Mean Opinion Score) test to further validate our designed function \cref{eq:score}. We sampled 100 artworks uniformly across the range of scores from the proposed BAID, and asked 10 college students majoring in art and design to score these samples. Results of the MOS test and more discussion of the score-generating function are given in Section 2 of the supplementary material. }

\section{Style-specific Art Assessment Network}
\label{sec:SAAN}
In this section, we introduce our proposed approach Style-specific Art Assessment Network (SAAN), which uses style-specific and generic aesthetic features to evaluate artistic images. 
SAAN consists of three modules: 1) the Style-specific Aesthetic Branch (SAB) extracts style-specific aesthetic features (Sec.~\ref{ssec:style-specific}); 2) the Generic Aesthetic Branch (GAB) extracts generic aesthetic features based on self-supervised learning (Sec.~\ref{sec:GAB}); and 3) the Spatial Information Fusion Module fuses the spatial information using a non-local block and incorporates the composition of artwork into the assessment (Sec.~\ref{ssec:spatial-fusion}). 
To train deep models to work better for aesthetic evaluation, we pretain the network by applying different manipulations, and training the network to classify manipulations and recognize manipulation intensity.
The overall architecture of SAAN is displayed in \cref{fig:model}.

\subsection{Style-specific Aesthetic Feature Extraction}
\label{ssec:style-specific}
Intuitively, let us consider an oil painting $p_{1}$ and a pencil sketch $p_{2}$. From the perspective of human perception, when evaluating $p_{1}$, we may take into account the use of color and the variation of brushstrokes. Instead, we may put more emphasis on the control of lines when we evaluate $p_{2}$. Thus, the objective of the style-specific aesthetic branch is to extract the aesthetic features of the given artwork appropriate to its artistic style.

Style representations have been heavily discussed and studied in the field of style transfer. However, to the best of our knowledge, none of the existing IAA methods has considered the integration of style information into the prediction model. We follow the mainstream style transfer approaches~\cite{huang2017arbitrary, li2017universal, park2019arbitrary, liu2021adaattn} to use an ImageNet \cite{deng2009imagenet} pretrained VGG-19 \cite{simonyan2014very} backbone $F_{sty}$ to extract style features. To extract \YL{aesthetics}-related features, we use a ResNet-50 \cite{he2016deep} backbone $F_{aes}$, which is pretrained by the self-supervised scheme in Sec.~\ref{section:pretrain}. Given an image $p$, the style features $f_{sty}$ and the aesthetic features $f_{aes}$ are extracted by:
\begin{equation}
\begin{aligned}
    f_{sty} &= F_{sty}(p, \theta_{sty}) \\
    f_{aes} &= F_{aes}(p, \theta_{aes}),
    \end{aligned}
\end{equation}
where $\theta_{sty}$ and $\theta_{aes}$ are the parameters of $F_{sty}$ and $F_{aes}$ respectively.

Instead of concatenating the style and aesthetic features together as input to subsequent network structures, we add an AdaIN \cite{huang2017arbitrary} layer to integrate style information in $f_{sty}$ into the aesthetic feature $f_{aes}$. Given a content feature map $x$ and a style feature map $y$, AdaIN encodes the content and style information in the feature space by aligning the channel-wise mean and variance of $x$ to match those of $y$:
\begin{equation}
    AdaIN(x,y) = \sigma(y)\cdot\frac{x-\mu(x)}{\sigma(x)} + \mu(y).
\end{equation}

Here we take the advantage that Huang \etal \cite{huang2017arbitrary} mentioned in their study: the output produced by AdaIN will have the same high average activation for the specific style feature, while preserving the spatial structure of the image.

The final output of the SAB is a style-specific aesthetic feature $f_{aes_{s}}$ calculated as follows:
\begin{equation}
    f_{aes_{s}} = AdaIN(f_{aes}, f_{sty}),
\end{equation}
\ie, the style-specific aesthetic feature $f_{aes_{s}}$ is obtained by changing the aesthetic feature $f_{aes}$ to incorporate the style information of $f_{sty}$.

\subsection{Generic Aesthetic Feature Extraction}
\label{sec:GAB}
In addition to the style-specific aesthetic branch, we propose a generic aesthetic branch to extract the aesthetic features shared by common categories of artworks. Aesthetic attributes like the integrity of the salient component and the layout of the frame can be viewed as intrinsic requirements.

We use 
ResNet-50 
as the backbone to extract generic aesthetic and apply a self-supervised scheme to pretrain the backbone. Simply put, the pretraining stage includes two pretext tasks, one is to classify the applied distortions and the other is to estimate the intensity of the applied distortions. See \cref{section:pretrain} for further details. Denote the backbone as $F_{gen}$, the output generic aesthetic feature $f_{aes_{g}}$ of a given image $p$ is obtained by:
\begin{equation}
    f_{aes_{g}} = F_{gen}(p, \theta_{gen}),
    \label{eq:gen}
\end{equation}
where $\theta_{gen}$ is the parameters of $F_{gen}$.

\subsection{Self-supervised Pretraining Scheme}
\label{section:pretrain}
Pfister \etal \cite{Pfister_2021_CVPR} argue that ImageNet-pretrained backbones are not well-suited to the IAA task. For instance, such classification model should be invariant to the image’s brightness and thus prohibits taking the image’s brightness into account when evaluating its aesthetics. Sheng \etal \cite{sheng2020revisiting} \YL{state} that a trained IAA model is able to distinguish fine-grained aesthetic differences caused by various image manipulations.
Based on the observation that certain distortions applied to images will reduce their appeal, both \YL{works} \cite{Pfister_2021_CVPR, sheng2020revisiting} proposed a self-supervised scheme to pretrain the backbone of \YL{an} IAA model.

\begin{figure}[t]
    \centering
    \includegraphics[width=1.\linewidth]{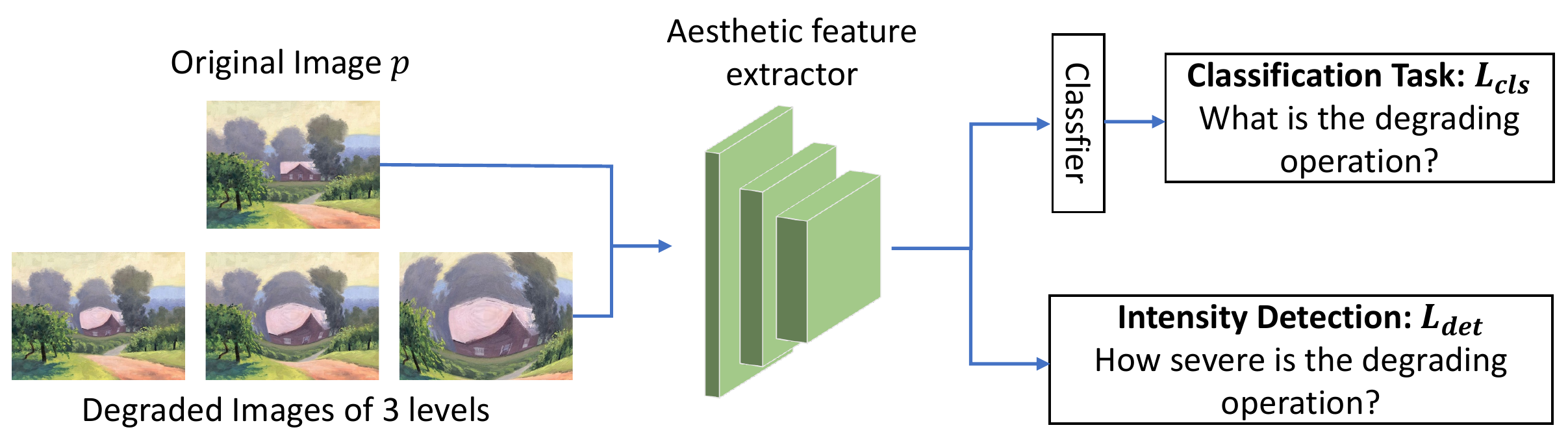}
    \caption{Pretraining pipeline. We first edit the original image using a distortion operation that reduces its appeal with three different levels. Then we train the aesthetic feature extractor with two pretext tasks: one to identify the type of the distortion, and the other to detect the intensity of the distortion.}
    \label{fig:pretrain}
\end{figure}

Inspired by these works, we adopt a pretraining approach similar to the one proposed in \cite{sheng2020revisiting} in our work which includes two aesthetics-aware pretext tasks: one to identify the type of the distortion applied to a given image; and the other to detect the intensity of the applied distortion.
The whole pretraining pipeline is shown in \cref{fig:pretrain}.


\textbf{Degradation editing operations.}
Based on the selection of manipulations in \cite{Pfister_2021_CVPR, sheng2020revisiting}, we carefully select a variety of image manipulation operations to reduce artistic appeal. We design operations with different parameters for generating artificial training instances, which are listed in \cref{tab:opt}.

There are two main differences between the operations and the parameters we choose and the ones used in \cite{sheng2020revisiting}:

\begin{itemize}
    \item The operation list used in \cite{sheng2020revisiting} ignores the distortion to some global aesthetic factors, \eg, rule of thirds. We add operations that distort the layout and the composition of the original image (\eg cropping, convex). We also add art-related distortions, \eg, stylization which generates unwanted lines given a fine artwork.
    \item \cite{sheng2020revisiting} only adopts two levels of distortions controlled by two sets of parameters. We carefully check the effect of the operations under different parameters and apply more subtle levels of distortion by using three sets of parameters (except for the rotation).
\end{itemize}

\begin{table}[t]
\caption{The operation list used in the \YL{pretraining} pipeline. Operations marked in \textcolor{red}{red} color are our newly added ones which are not included in the operation list in \cite{sheng2020revisiting}.}
\resizebox{.95\columnwidth}{!}{
\begin{tabular}{cc}
\toprule[2pt]
\textbf{Manipulation} & \textbf{Parameter}                 \\ \midrule[1pt]
Gaussian noise        & 0.2, 0.4, 0.8                           \\
Quantization          & 64, 32, 8                  \\
Gaussian Blur         & 0.4, 0.8, 2                          \\
Exposure              & 1.5, 2.0, 2.5                           \\
Rotation              & 45, -45                            \\
\textcolor{red}{Cropping}           &  3/4, 2/3, 1/2                             \\
\textcolor{red}{Stylization}       & (50, 0.6), (50, 0.3), (50, 0.1)           \\
\textcolor{red}{Convex}                & 1/8, 1/4, 1/2                           \\
\textcolor{red}{PencilSketch}      & (100, 0.1, 0.02), (100, 0.4, 0.02), (100, 0.6, 0.02) \\
\textcolor{red}{CutMix} \cite{yun2019cutmix}                & 32, 64, 128                             \\
None                  &  -\\
\bottomrule[2pt]
\end{tabular}}

\label{tab:opt}
\end{table}

\begin{figure}[t]
  \centering
   \includegraphics[width=0.95\linewidth]{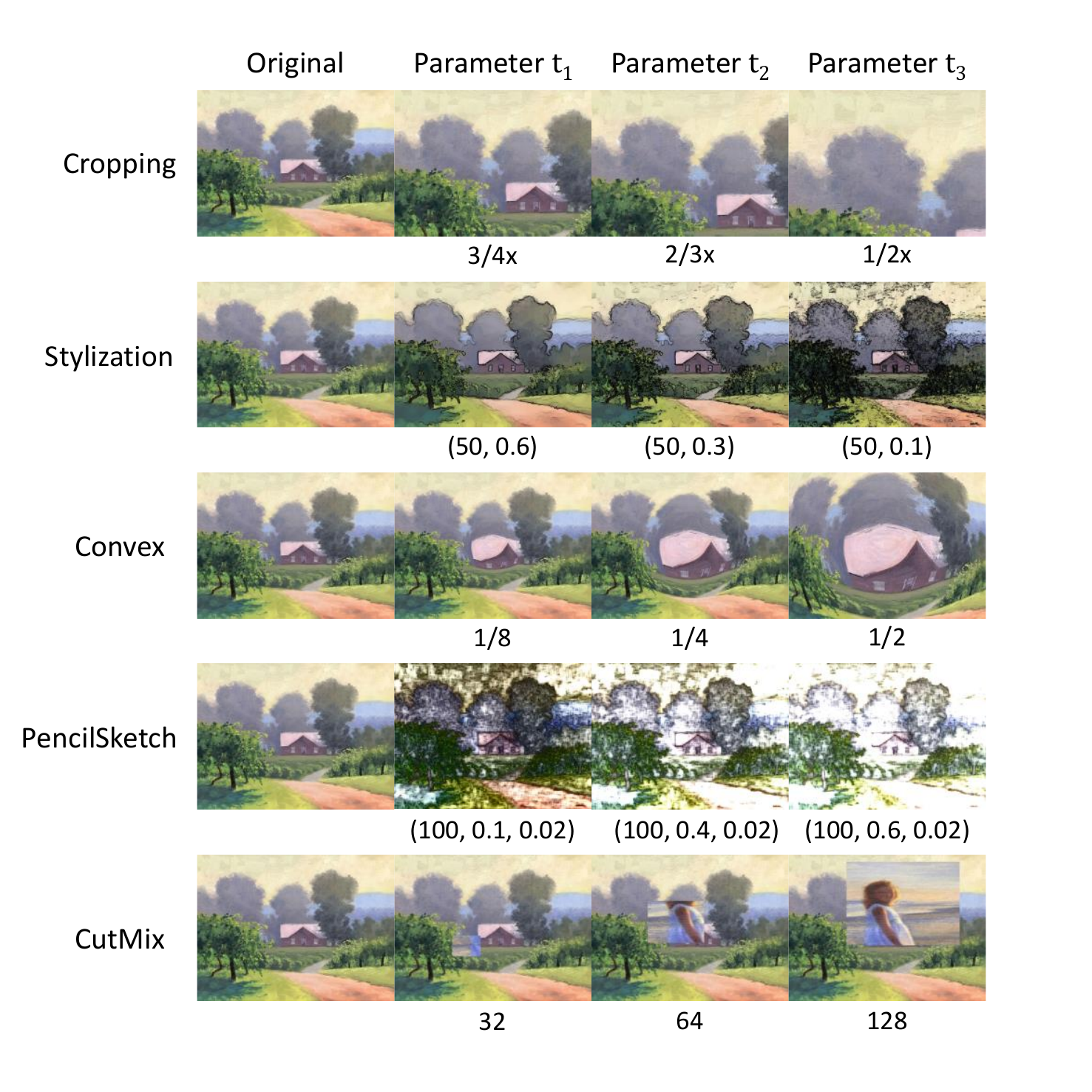}
   \caption{Visualization of the editing effects of the newly added operations in \cref{tab:opt}.}
   \label{fig:manipulations}
\end{figure}

\textbf{Distortion classification pretext task.} This task is identical to the classification task proposed in \cite{sheng2020revisiting, Pfister_2021_CVPR}. Denote an image
patch as $p$, the loss term of the classification task $L_{cls}(p, t)$ reinforces the model to recognize which operation $t$ has been
applied to $p$:
\begin{equation}
\begin{aligned}
    L_{cls}(p, \theta_{t}) &=  -\log(P_{t}(p;W)) \\
    P_{t}(p;W) &= P(\hat{t} = t| m(p, \theta_{t});W)
\end{aligned}
\end{equation}
where $m(p, \theta_{t})$ is the manipulated
output patch given the image patch $p$ by the parameters $\theta_{t}$, and $P(\hat{t}= t| m(p, \theta_{t});W)$ is the probability predicted by our model $W$ that $p$ has undergone a degradation operation of type $\hat{t}$ \YL{that matches ground truth operation $t$}. 
Note that \YL{for this task,} operations of different parameters are viewed as different operations.

\textbf{Intensity detection pretext task.} To detect the intensity of the distortion, Sheng \etal \cite{sheng2020revisiting} proposed a triplet loss $L_{trp}$, which enforces a smaller distance between original patch and a slightly distorted patch, and a larger distance between original patch and a highly distorted patch. Given operation $t$ and two control parameters $\theta_{t_{1}}$ and $\theta_{t_{2}}$, $L_{trp}$ is calculated using \cref{eq:triplet}:

\begin{small}
\begin{equation}
\begin{aligned}
D(p,\theta_{t}) &= ||h(p,W)-h(m(p,\theta_{t}),W)||_2^2 \\
L_{trp}(p,\theta_{t_{1}},\theta_{t_{2}})&=max\{0, 1+D(p,\theta_{t_{1}})-D(p,\theta_{t_{2}})\}
\end{aligned}
\label{eq:triplet}
\end{equation}
\end{small}
where $h(p,W)$ is the L2-normalized feature 
\YL{of patch $p$}
extracted from the model with \YL{parameters} $W$, \YL{and $D(p,\theta_t)$ works out the squared difference between the normalized features before and after applying the manipulation.}

However, we find that two levels of distortion will make the $L_{trp}$ hard to converge in our experiment. Meanwhile, the difference between the downgrading effect of $\theta_{t_{1}}$ and $\theta_{t_{2}}$ are sometimes too large, \eg, Gaussion noise with $\theta_{t_{1}}=0.2$ and $\theta_{t_{2}}=0.8$ is not a smooth intensity transition. Thus, we apply {\bf three levels of distortion} $\theta_{t_{1}}$, $\theta_{t_{2}}$, $\theta_{t_{3}}$ and introduce $L_{det}$ as:

\begin{equation}
L_{det} = L_{trp}(p,\theta_{t_{1}},\theta_{t_{2}}) + L_{trp}(p,\theta_{t_{2}}, \theta_{t_{3}}) - 1
\label{eq:3level}
\end{equation}

The overall loss of the pretraining pipeline is:
\begin{equation}
    L = L_{cls} + \lambda L_{det},
\end{equation}
where $\lambda$ is used to balance the two terms.

\subsection{Spatial Information Fusion}
\label{ssec:spatial-fusion}
Previous works \cite{she2021hierarchical, hosu2019effective} have demonstrated that the layout of the given image is critical when predicting its aesthetic score. In this work, we add a non-local block \cite{wang2018non} before the Multi-Layer Perception (MLP) to fuse the spatial information and implicitly detect the composition of the artwork. Specifically, given the extracted features $f_{aes_{s}}$ and $f_{aes_{g}}$ from the two branches mentioned above, the features are passed through one non-local block $F_{nlb}$ after proper resizing and concatenation:
\begin{equation}
    f_{out} = F_{nlb}(f_{aes_{s}} \oplus f_{aes_{g}}, \theta_{nlb}),
\end{equation}
where $\oplus$ denotes the concatenate operation, and $\theta_{nlb}$ is the parameters of the non-local block.

Finally, \YL{an} MLP $L$ is used to output the predicted score $s_{pred}$ of the input image $p$:
\begin{equation}
    s_{pred} = L(f_{out}, \theta_{L}).
\end{equation}

\section{Experiments}
In this section, we first describe the experiment settings used in the pretraining pipeline (introduced in \cref{section:pretrain}), then we elaborate the training and evaluation of the proposed SAAN and state-of-the-art IAA methods on BAID, and conduct ablation studies to validate the effectiveness of each module.

\subsection{Experimental Setup}

\noindent \textbf{Pretraining Settings.} 
Instead of using a subset of ImageNet for pretraining~\cite{sheng2020revisiting}, we directly use our BAID as the pretraining dataset \HY{to meet the requirements mentioned in \cite{Pfister_2021_CVPR}}. We adopt the pretraining pipeline to train the ResNet-50 aesthetic feature extractor. 
For each image in BAID, we randomly choose three manipulation operations in \cref{tab:opt} to edit it. 
We apply the Adam optimizer using a batch size of 64, with the weight decay of $5e-4$. We begin with a learning rate of $1e-3$, dropped it by a factor of 0.1 after every 10 epochs. Following the settings in \cite{sheng2020revisiting}, we activate $L_{det}$ with $\lambda = 0.1$ after the first 30 epochs.

\noindent \textbf{Training Settings.} After pretraining, we then train the overall pipeline using the mean squared error (MSE) loss between predicted and ground truth aesthetic scores. Previous experiments on IAA \cite{murray2012ava, kong2016photo} have reported that inappropriate data augmentation during training will degrade the performance at test time. Therefore, in the training stage, we directly resize the original image to 224 $\times$ 224 to avoid cropping which may decrease the aesthetic quality. We apply \YL{the} Adam \YL{optimizer} using a batch size of 64. We begin with a learning rate of $1e-5$, dropped it by a factor of 0.1 every 10 epochs for the first 40 epochs. Following the settings in \cite{huang2017arbitrary}, we freeze the VGG backbone in the style-specific aesthetic branch, and further freeze the ResNet-50 backbone in the generic aesthetic branch to avoid overfitting into a certain category of artistic style. We randomly split the 60,337 images in BAID into 53,937:6,400 for training and testing respectively.

\noindent \textbf{Evaluation Metrics.} Typically, IAA methods are evaluated on regression and classification tasks. To evaluate the regression performance, we adopt two popular evaluation metrics: 1) Spearman’s rank correlation coefficient (SRCC) $S$ and 2) Pearson correlation coefficient (PCC) $P$. We also convert the predicted and ground-truth scores to binary-class labels (attractive \& unattractive art) using a threshold of 5 \YL{(midpoint from 0 to 10)}
and calculate the accuracy. 

\begin{table}[t]
\caption{Comparison with state-of-the-art open-source IAA methods on BAID.}
\resizebox{.95\columnwidth}{!}{
\begin{tabular}{lccccc}
\toprule[2pt]
\textbf{Methods} &\textbf{\#Params} &\textbf{SRCC $\uparrow$ } & \textbf{PCC $\uparrow$} & \textbf{Accuracy $\uparrow$} \\ \midrule[1pt]
NIMA \cite{talebi2018nima} & 63.61M & 0.393         & 0.382        & 71.01\%           \\
${\rm MP}_{ada}$ \cite{sheng2018attention} & 63.37M &  0.437         & 0.425        & 74.33\%           \\
MLSP    \cite{hosu2019effective} & 73.97M & 0.441         & 0.430        & 74.92\%           \\
BIAA     \cite{zhu2020personalized}  &   97.49M   & 0.389         & 0.376        & 71.61\%           \\
TANet  \cite{ijcai2022p132}  &   57.87M    & 0.453         & 0.437        & 75.45\%           \\
\hline
\textbf{Ours}  & 64.44M    & \textbf{0.473}         & \textbf{0.467}        & \textbf{76.80\%}       \\ \bottomrule[2pt]    
\end{tabular}}
\label{tab:eval}
\end{table}

\subsection{Performance Comparison}
\label{ssec: performance}
We compare our method with five state-of-the-art open-source IAA methods on our BAID dataset, including NIMA \cite{talebi2018nima}, ${\rm MP}_{ada}$ \cite{sheng2018attention}, MLSP \cite{hosu2019effective}, BIAA \cite{zhu2020personalized} and TANet \cite{ijcai2022p132}. 
Note that most IAA methods are trained using EMD loss, which requires ground truth score distributions rather than only mean scores for training. 
Therefore, we modify the code provided by the researchers and make them trainable on BAID, which accounts for the reason that we only compare SAAN with open-source methods.

\cref{tab:eval} shows the performance of the IAA methods and our SAAN on BAID. Compared with these methods, our SAAN model achieves the best performance on all metrics. This suggests that understanding the style of the artistic image assists in perceiving the aesthetics of the image, especially when there are a wide variety of styles that may have different evaluation criteria. \HY{For more performance evaluation, please refer to Section 3 of the supplementary material.}

\begin{table}[t]
\caption{Ablation study results on the BAID.}
\resizebox{.95\columnwidth}{!}{
\begin{tabular}{lccc}
\toprule[2pt]
\textbf{Method}            & \textbf{SRCC $\uparrow$} & \textbf{PCC $\uparrow$} & \textbf{Accuracy $\uparrow$} \\ \midrule[1pt]
w/o style-specific branch  & 0.425         & 0.411        & 73.22\%           \\
\HY{w/o generic aesthetic branch}  & 0.439         & 0.426        & 74.60\%         \\
w/o new editing operations & 0.460         & 0.445        & 76.14\%           \\
w/o 3-level \YL{manipulation}               & 0.462         & 0.448        & 76.19\%           \\
w/o spatial information fusion              & 0.459         & 0.440        & 76.14\%           \\
\hline
\textbf{Ours }                      & \textbf{0.473}         & \textbf{0.467}        & \textbf{76.80\%}          \\ \bottomrule[2pt]  
\end{tabular}}

\label{tab:ablation}
\end{table}

\subsection{Ablation Study}
\cref{tab:ablation} shows the ablation study results. (1) We first examine the effectiveness of the style-specific branch \HY{and the generic branch}. All three metrics drop drastically when SAB is removed, where SRCC drops from 0.473 to 0.425, PCC drops from 0.467 to 0.411 and Accuracy drops from 76.80\% to 73.22\%. \HY{After removing the generic aesthetic branch, SRCC drops from 0.473 to
0.439, PCC drops from 0.467 to 0.426 and Accuracy drops
from 76.80\% to 74.60\%.} The disparity indicates that incorporating the style information with the generic information is of great help when evaluating artworks. 
(2) We then compare our new operation list (\cref{tab:opt}) and the one proposed in \cite{sheng2020revisiting}. SAAN pretrained using our \YL{proposed} list gives better \YL{results}, which shows that adding global and art-related manipulations makes the model fit better \YL{in} the artistic field. (3) We further analyze the efficacy of adopting 3 levels of distortions by deleting a set of parameters during pretraining, \ie, using 2 levels. Results demonstrate that a more fine-grained intensity setting benefits the model to learn \YL{aesthetics}-related features. (4) Finally, we remove the spatial information fusion module of the SAAN framework \YL{and the results demonstrate} that fusing the spatial information \YL{enhances} the performance of AIAA models. 

\section{Conclusions}
This paper addresses the challenging task of artistic image aesthetic assessment (AIAA). To achieve this goal, we create a large-scale dataset BAID, which is constructed completely from artworks, including 60,337 artworks annotated with more than 360,000 votes. BAID is, to our knowledge, the largest artistic image aesthetic assessment dataset, and far exceeds existing IAA and AIAA datasets in quantity and quality of artworks.
We further set up a complete benchmark and develop a baseline model called SAAN,  
which introduces adaptive perception to extract style-specific aesthetic features and achieves state-of-the-art performance on the proposed dataset.
We hope our contributions will motivate the community to rethink AIAA and stimulate research with a broader perspective.

~\\
 \noindent
\textbf{Acknowledgements.} This work was supported by National Natural Science Foundation of China (72192821, 61972157, 62272447), Shanghai Municipal Science and Technology Major Project (2021SHZDZX0102), Shanghai Science and Technology Commission (21511101200), Shanghai Sailing Program (22YF1420300, 23YF1410500), CCF-Tencent Open Research Fund (RAGR20220121) and Young Elite Scientists Sponsorship Program by CAST (2022QNRC001).


\clearpage
\renewcommand\thesection{\Alph{section}}
\setcounter{section}{0}

\noindent
\textbf{\Large Appendix}
\section{Overview}
In this supplementary material, more discussion, visualization and experimental results are provided, which are organized as follows:
\begin{itemize}
    \item \cref{sec:boldbrush} provides more details about the data collection (\cref{subsec:user}) and analysis (\cref{subsec:ana}) of the proposed BAID dataset \HY{and the results of the MOS (Mean Opinion Score) test mentioned in the main paper (\cref{subsec:MOS})}.
    \item \cref{sec:ablation} conducts an ablation study on each of our newly added operations.
    \item \cref{sec:results} provides an evaluation of the style-specific aesthetic branch in the proposed SAAN (\cref{subsec:style}), \HY{evaluates the performance of SAAN on the AVA dataset \cite{murray2012ava} (\cref{subsec:datasets})}, and gives more prediction results on the test set of BAID (\cref{subsec:vis}).
\end{itemize}

\section{More analysis of the proposed BAID}
\label{sec:boldbrush}
\subsection{User Interface of BoldBrush}
\label{subsec:user}
In Section 3 of the main paper, we discuss the construction of our proposed BoldBrush Artistic Image Dataset (BAID). \cref{fig:bb} shows the detail page of an entry on the BoldBrush (\url{https://faso.com/boldbrush/popular}) website:

\begin{figure}[htbp]
    \centering
    \includegraphics[width=.9\linewidth]{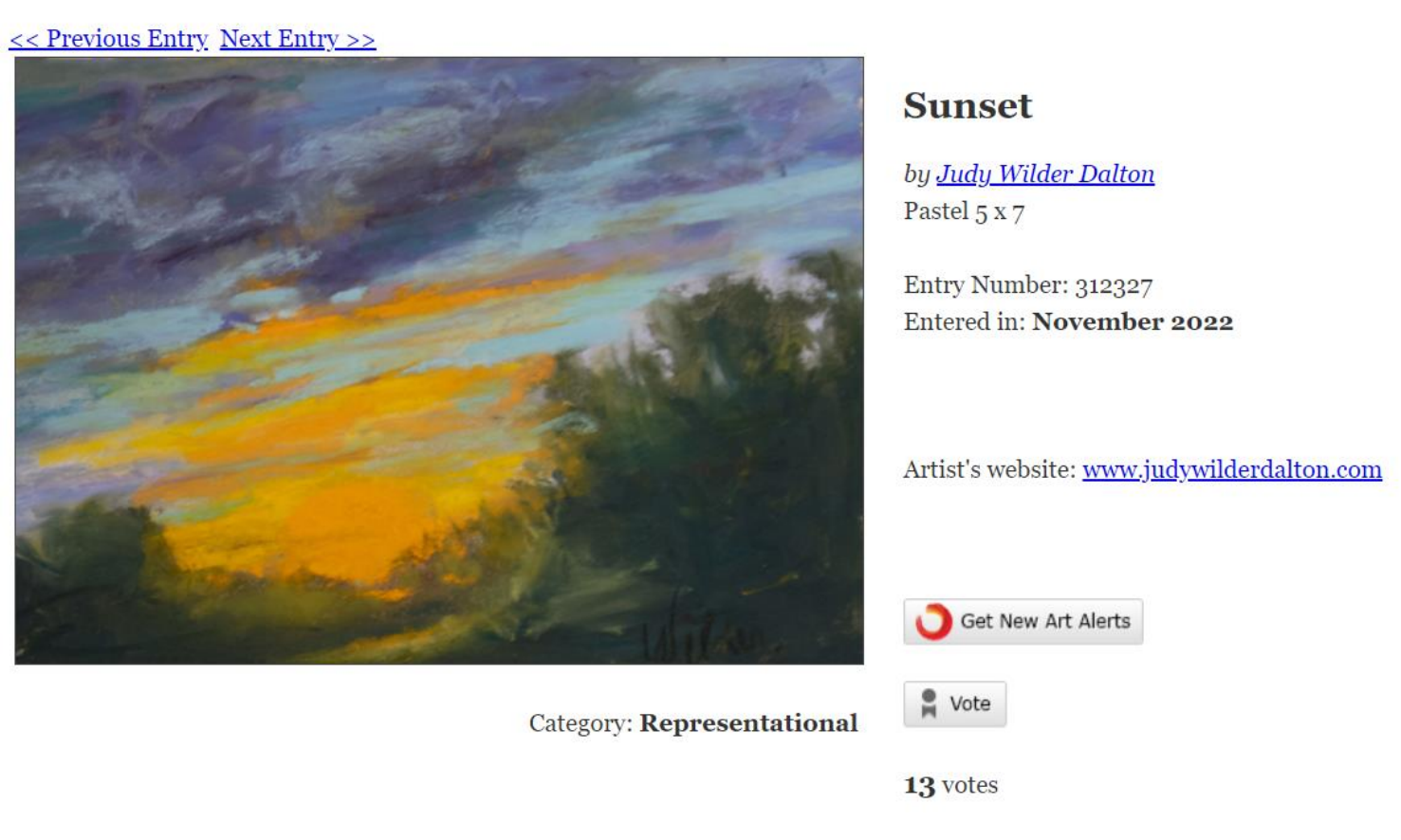}
    \caption{Interface of the BoldBrush website}
    \label{fig:bb}
\end{figure}

The available information of an entry on BoldBrush includes: the title; the artist; the painting medium; the entry number and month of entry; the number of votes; the category of the entry. In this work, we utilize the number of votes and the month of entry to generate score annotations and form a large-scale artistic image aesthetic assessment dataset, BAID. Meanwhile, we collect and save all the above information. We believe the BAID dataset can effectively serve as a foundation for constructing artistic image datasets for other purposes, \eg, developing automatic artist \cite{cetinic2013automated, david2016deeppainter} and style \cite{shamir2010impressionism, bar2014classification} classification methods.

\begin{table}[t]
\centering
\caption{The most frequently used painting media in BAID and the average score of artworks created in these media.}
\setlength{\tabcolsep}{0.5mm}{
\begin{tabular}{ccc}
\toprule[2pt]
\textbf{Painting Medium}  & \textbf{Number of Images} & \textbf{Average Score $\uparrow$} \\ \midrule[1pt]
Oil  & 38,586 & 4.27\\
Acrylic & 6,733  & 4.30\\
Watercolor & 5,328 & 4.24\\
Pastel & 5,156 & 4.22\\
Pencil & 1,063 & 4.34\\\bottomrule[2pt]
\end{tabular}}
\label{tab:medium}
\end{table}

\begin{table}[t]
\centering
\caption{Correlation between scores and hand-crafted features.}
\setlength{\tabcolsep}{12mm}{
\begin{tabular}{cc}
\toprule[2pt]
\textbf{Features}  & \textbf{SRCC $\uparrow$ } \\ \midrule[1pt]
Colorfulness  & 0.011 \\
Contrast & 0.049  \\
Sharpness & 0.029 \\
Complexity & 0.014 \\\bottomrule[2pt]
\end{tabular}}
\label{tab:cor}
\end{table}

\subsection{Further analysis}
\label{subsec:ana}
\HY{The generation of the score annotations in BAID is based on votes, which makes it hard to filter out unreasonable labels. To eliminate the concern about bias,} we calculate the most frequently used painting media and the average scores of the artworks created using the specific media. The results are shown in \cref{tab:medium}. Furthermore, following the benchmarks applied in \cite{mould2017developing}, we calculate several hand-crafted features and measure their correlation (\ie, the Spearman Rank-order Correlation Coefficient, SRCC) with the scores of the artworks in the BAID dataset. We randomly select 6,400 images from BAID and the results are shown in \cref{tab:cor}. The results indicate that the proposed BAID suffers little from art preference bias and is of high credibility.

\HY{There is a potential concern regarding the data imbalance mentioned in Section 3.3 of the main paper. The score distribution of BAID is imbalanced but it reflects the realistic distribution. We did consider reducing the imbalance. However, the most effective way would be to abandon most of the images with low votes, which would result in a significant drop in the size of BAID. Besides, the imbalance is related to the nature of the original data, and we believe
that a well-developed IAA method should be able to deal with
such an imbalance.}

\subsection{Results of the MOS test}
\label{subsec:MOS}
\HY{As mentioned in Section 3.3 of the main paper, we sampled 100 artworks uniformly across the range of scores from the proposed BAID. We asked 10 college students majoring in art and design to score for these samples and calculated the mean opinion score (MOS)
for each sample. We compared several designed functions we have experimented with during the construction of BAID. In the following equations, $v_{i}$ denotes the number of votes of the image, $\bar{v}_{m_{i}}$ denotes the average number of votes of the month $m_{i}$, $\hat{v}_{m_{i}}$ denotes the maximum number of votes of the month $m_{i}$, and $s_{i}$ denotes the generated score.

Choice A:
\begin{equation}
    s_{i} = 5 \times \frac{v}{\bar{v}_{m_{i}}},
\label{eq:scoreA}
\end{equation}

Choice B:
\begin{equation}
\begin{aligned}
    s_{i} &= 5 - 5 \times \frac{\bar{v}_{m_{i}}-v}{\bar{v}_{m_{i}}}, (v \leq \bar{v}_{m_{i}}) \\
    s_{i} &= 5 + 5 \times \frac{v}{\hat{v}_{m_{i}} - \bar{v}_{m_{i}}}, (v > \bar{v}_{m_{i}})
\end{aligned}
\label{eq:scoreB}
\end{equation}

Choice C:
\begin{equation}
\begin{aligned}
    s_{i} &= 5 - v \times \frac{\bar{v}_{m_{i}}-v}{\bar{v}_{m_{i}}}, (v \leq \bar{v}_{m_{i}}) \\
    s_{i} &= 5 + 5 \times \frac{v}{\hat{v}_{m_{i}} - \bar{v}_{m_{i}}}, (v > \bar{v}_{m_{i}})
\end{aligned}
\label{eq:scoreC}
\end{equation}

Ours:
\begin{equation}
\begin{aligned}
    x_{i} &= \frac{\bar{v}_{m_{i}} - v_{i}}{\bar{v}_{m_{i}}},\\
    s_{i} &= 10 \times \frac{1}{1+e^{x_{i}}}  ,
\end{aligned}
\label{eq:scoreD}
\end{equation}

\begin{table}[htbp]
\centering
\caption{Comparison of different score-generating functions}
\begin{tabular}{lccc}
\toprule[2pt]
\textbf{Method}      & \textbf{SRCC $\uparrow$}  & \textbf{RMSE $\downarrow$}  \\
\hline
A           & 0.221 & 0.980 \\
B           & 0.576 & 0.502 \\
C           & 0.594 & 0.492 \\
\hline
\textbf{Ours} & \textbf{0.734} & \textbf{0.305} \\
\bottomrule[2pt]  
\end{tabular}
\label{tab:MOS}
\end{table}

Note that, images with $\bar{v}_{m_{i}}$ votes are supposed to be given the score of 5, which leaves us few options when designing the score-generating function. We calculated the spearman rank-order correlation coefficient (SRCC) and root mean squared error (RMSE) between the scores generated by the above functions and the MOS results. The results are shown in \cref{tab:MOS}, which indicates that our chosen method better reflects human aesthetics.

The designed method seems similar to and may be confused with psychometric scaling of human votes\cite{mikhailiuk2018psychometric}. However, the votes in BAID are different from the ones commonly used in psychometric scaling tasks since a vote itself is not a personal opinion score or a binary variable.}

\section{More ablation study results}
\label{sec:ablation}
In Sections 4 and 5 of the main paper, we demonstrate the effectiveness of our proposed operation list compared to the one in \cite{sheng2020revisiting}. Here we provide more results of the ablation study on each of the newly added operations:

\begin{table}[htbp]
\centering
\caption{Ablation study results on the newly added operations.}
\resizebox{.9\columnwidth}{!}{
\begin{tabular}{lccc}
\toprule[2pt]
\textbf{Method}            & \textbf{SRCC $\uparrow$} & \textbf{PCC $\uparrow$} & \textbf{Accuracy $\uparrow$} \\ \midrule[1pt]
w/o cropping  &     0.471     &     0.463     &     76.59\%     \\
w/o stylization &    0.471      &     0.462     &   76.58\%     \\
w/o convex      &     0.471     &     0.464    &      76.60\%       \\
w/o pencilsketch  &     0.472     &  0.465       &     76.63\%       \\
w/o cutmix \cite{yun2019cutmix}  &     0.470     &    0.462     &   76.65\%  \\
w/o new editing operations & 0.460         & 0.445        & 76.14\%           \\
\hline
\textbf{Ours }                      & \textbf{0.473}         & \textbf{0.467}        & \textbf{76.80\%}          \\ \bottomrule[2pt]  
\end{tabular}}
\label{tab:ablation2}
\end{table}

We select one of the newly added operations at a time and discard it during the pretraining stage. The impact on the final assessment performance is shown in \cref{tab:ablation2}. Results demonstrate that all newly added operations improve the performance, where operations related to global aesthetic features (\eg Cutmix \cite{yun2019cutmix}, Cropping) are relatively more influential in learning aesthetic-aware features, while the PencilSketch operation is less powerful since it may generate low-level artifacts (\ie, unnecessary lines) and can trick the network into learning trivial features. We also experiment with all new editing operations removed, and it leads to more significant performance drop.

\section{More performance evaluation results}
\label{sec:results}

\subsection{Evaluation of the Style-specific Aesthetic Branch}
\label{subsec:style}
In Section 4.1 of the main paper, we propose a style-specific aesthetic branch, which adopts a VGG-19 \cite{simonyan2014very} backbone to extract the style feature of the input image, and incorporate the style information into aesthetic features to obtain style-specific aesthetic features. 

\begin{figure}[htbp]
\centering
    \includegraphics[width=.95\linewidth]{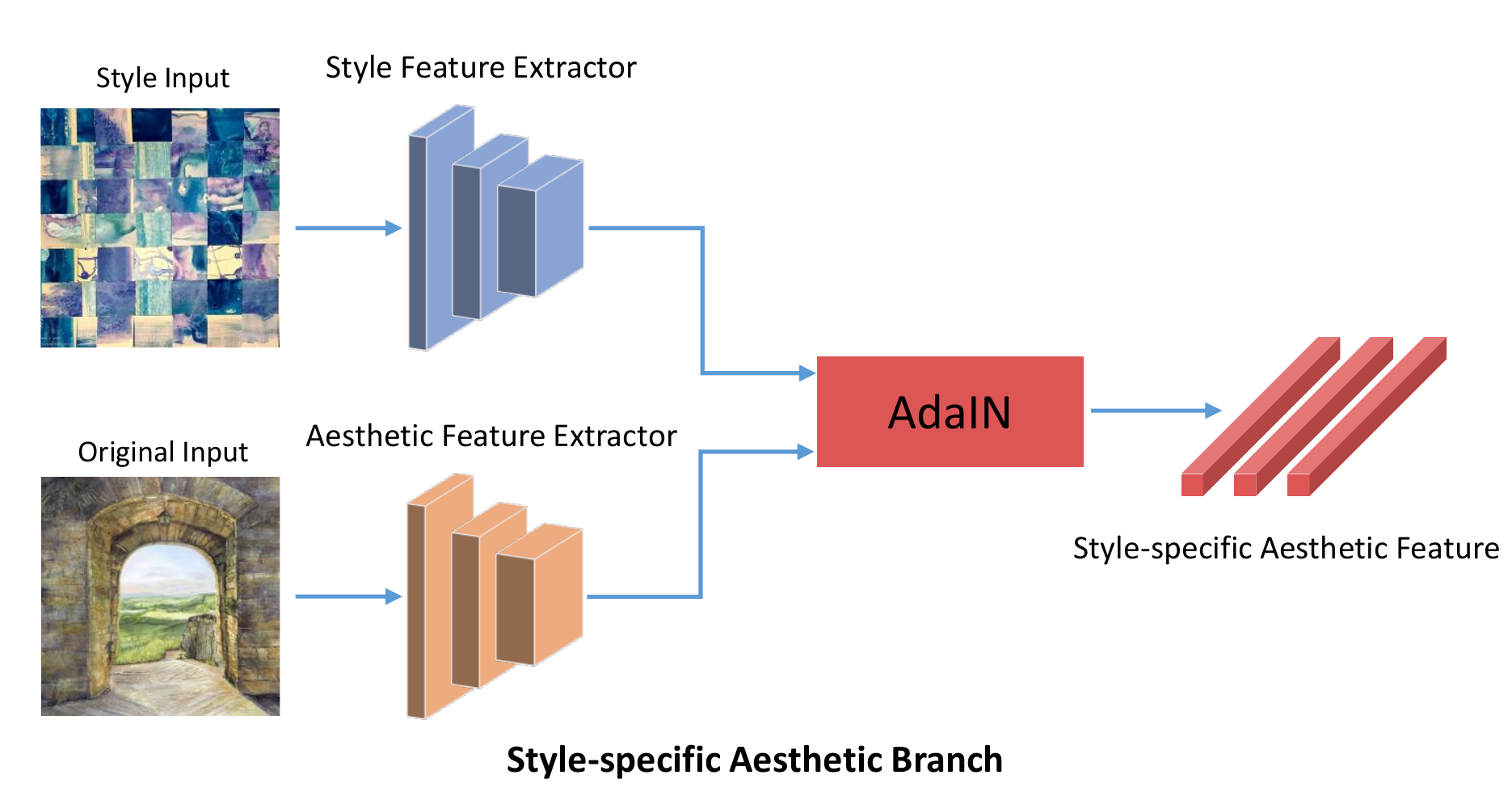}
    \caption{Validation of the style-specific aesthetic branch. We manually change the input to the style feature extractor (VGG-19) to be different from the original input in style. Note that the goal is still to predict the aesthetic score of the original input.}
    \label{fig:styleTest}
\end{figure}

To better illustrate the effect of incorporating style feature into the assessing process, given an artwork, we randomly select several images with different styles from the artworks in the test set of the BAID dataset, make them the input of the style feature extractor (VGG-19 backbone) and compare the predicted aesthetic scores. The experimental setting is shown in \cref{fig:styleTest}.
Since the goal of this branch is to extract style-related aesthetic features, if we extract a different style's feature, and incorporate the `wrong' style into the aesthetic features, then the calculated style-specific aesthetic feature is not dedicated to the current style, and the predicted aesthetic score is expected to decrease.

\begin{figure*}[t]
    \centering
    \includegraphics[width=.9\linewidth]{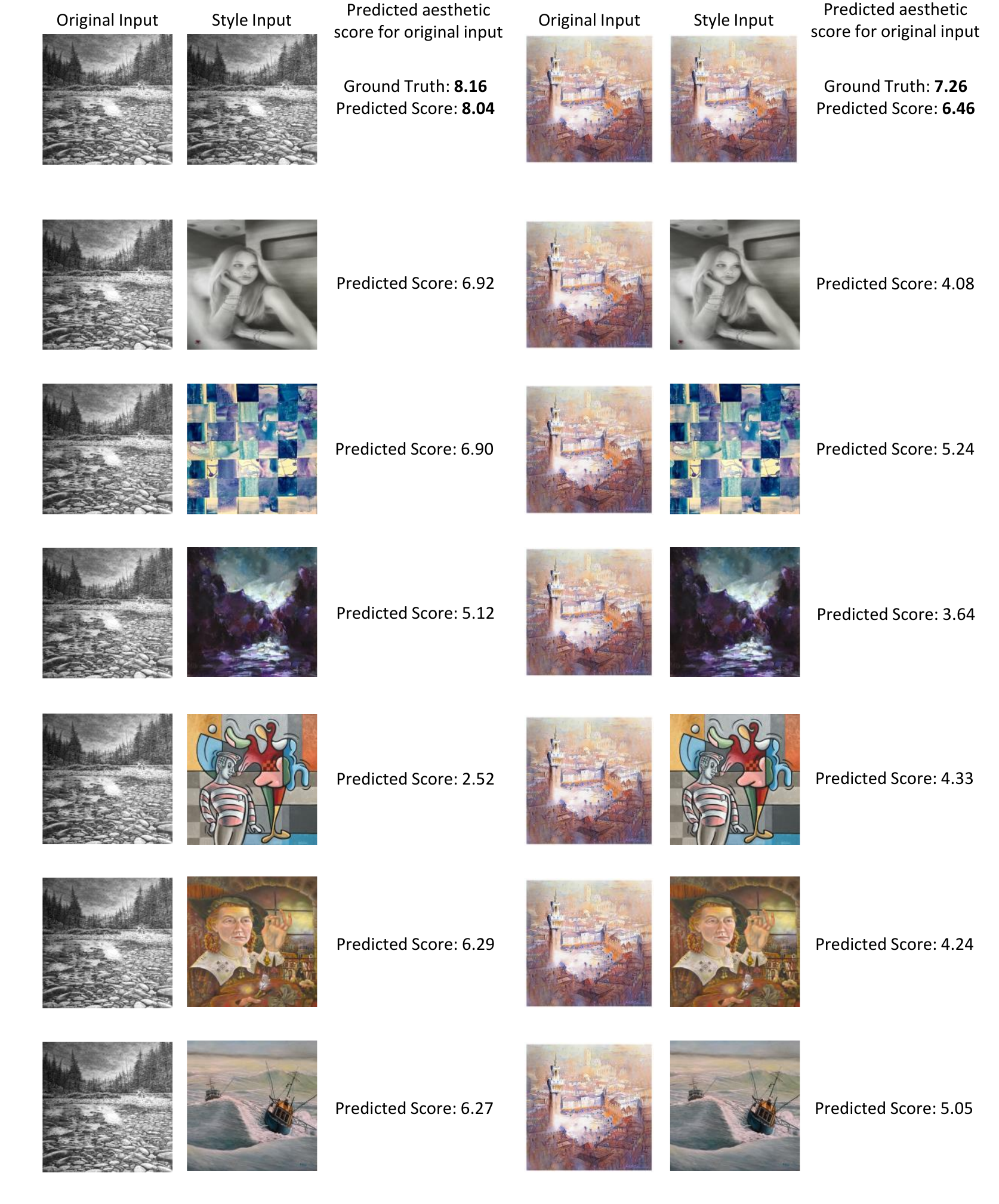}
    \caption{Prediction results of changing the input to the style feature extractor to be different from the original input in style (\cref{fig:styleTest}).}
    \label{fig:style_score}
\end{figure*}

\cref{fig:style_score} shows the results of using different style inputs when evaluating artistic images. When the style feature is extracted from an artwork with a different style from the original input, the predicted aesthetic score will decrease and the prediction error will increase, which further validates our idea of utilizing style information in the AIAA task.

\subsection{Performance on AVA dataset}
\label{subsec:datasets}
\HY{We modified the output layer of SAAN and trained it on AVA dataset\cite{murray2012ava} using EMD (Earth Mover's Distance) loss\cite{talebi2018nima}. \cref{tab:AVA} shows the performance of the state-of-the-art methods and SAAN on AVA dataset. The results of the state-of-the-art (SOTA) IAA methods come from the original papers and \cite{ijcai2022p132}. Our model gives competitive results compared with the SOTA methods. We believe SAAN works better on BAID since the distortions we used in the pretraining stage and the style-specific aesthetic branch are designed for and better suited to artistic images.}

\begin{table}[tbp]
\caption{Comparison with the SOTA IAA methods on AVA.}
\resizebox{.95\columnwidth}{!}{
\begin{tabular}{lccccc}
\toprule[2pt]
\textbf{Methods} & \textbf{SRCC $\uparrow$ } & \textbf{LCC $\uparrow$} & \textbf{Accuracy $\uparrow$} & \textbf{EMD $\downarrow$}\\ \midrule[1pt]
NIMA \cite{talebi2018nima} & 0.612 & 0.636 &  81.5\% &  0.050 \\
${\rm MP}_{ada}$ \cite{sheng2018attention} & 0.727 & 0.731 &  83.0\%  &  - \\
MLSP \cite{hosu2019effective} & 0.756 & 0.757 & 81.7\% &  \\
BIAA\cite{zhu2020personalized}  & 0.651 & 0.668 & - & - \\
PA\_IAA \cite{li2020personality} & 0.677 & - & 83.7\% & 0.049 \\
HLA-GCN\cite{she2021hierarchical} & 0.665 & 0.687 & \textbf{84.6\%} & \textbf{0.043} \\
TANet  \cite{ijcai2022p132} & \textbf{0.758} &  \textbf{0.765} & - & 0.047 \\
\hline
\textbf{Ours} & 0.742 &  0.748 & 80.6\% &  0.048 \\ \bottomrule[2pt]    
\end{tabular}}
\label{tab:AVA}
\end{table}

\subsection{Visualization of the prediction results}
\label{subsec:vis}
\begin{figure*}[t]
    \centering
    \includegraphics[width=.95\linewidth]{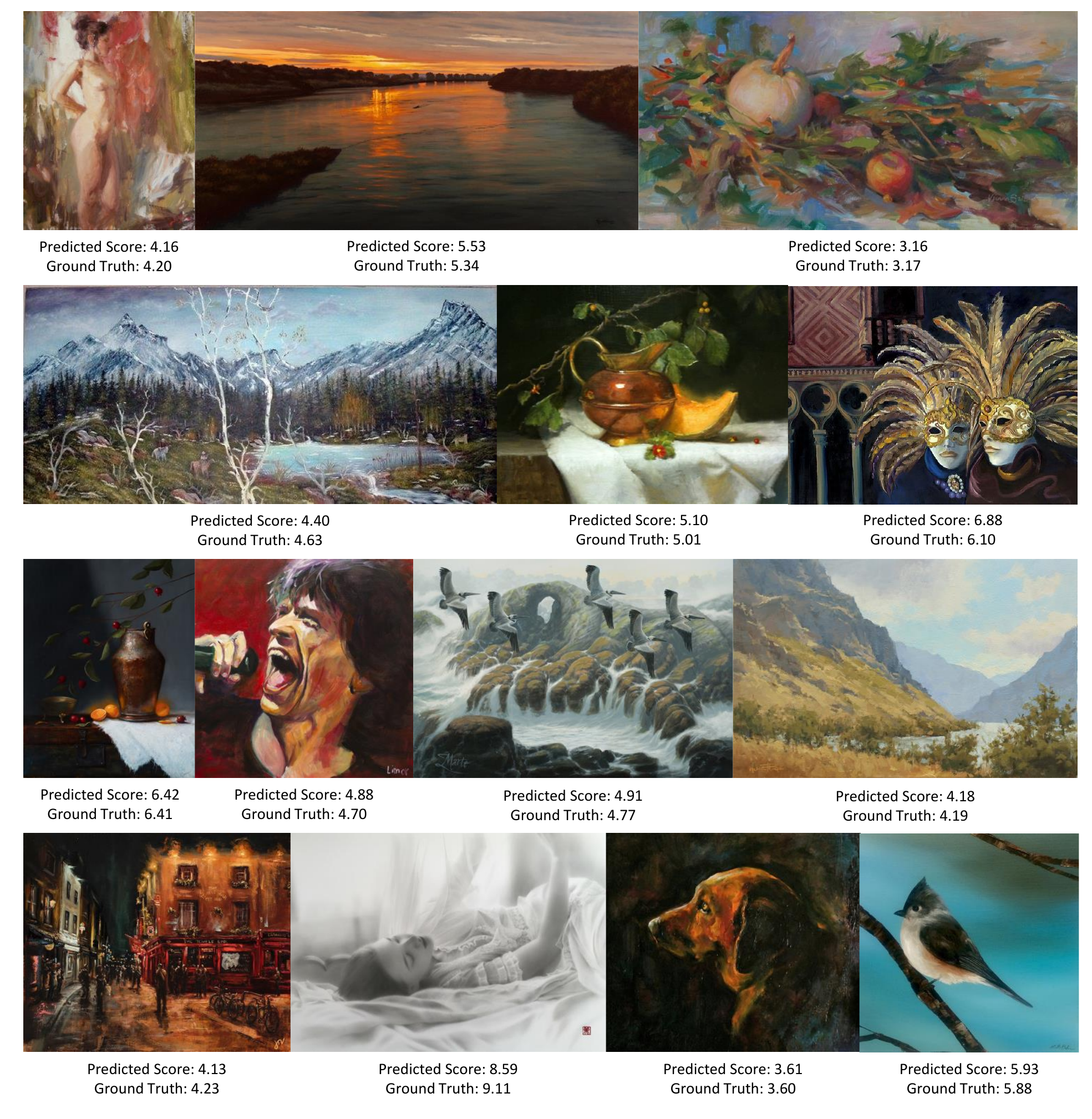}
    \caption{Some results on the test set of BAID, showing both the predicted scores by our method and ground truth scores.}
    \label{fig:testset}
\end{figure*}

\cref{fig:testset} shows the aesthetic score prediction results on some randomly picked artistic images from the test set of the proposed BAID dataset.

\end{document}